\documentclass[journal]{IEEEtran}

\usepackage{caption,subcaption,float}
\usepackage{graphicx}
\usepackage{multirow}
\usepackage{amsmath,fixmath,amsfonts}
\usepackage{array}
\usepackage{cite}
\usepackage[letterpaper]{geometry}
\usepackage{tikz}
\usepackage{enumitem}
\usepackage{tabularx}
\usepackage{array, multirow, bigdelim, makecell, booktabs} 
\usepackage{pgfplots}
\usepackage{framed,multirow, array}
\usepackage{comment}
\usepackage{graphicx,scalerel,stackengine}
\usepackage[numbers]{natbib}     
\usepackage{caption}
\usepackage{subcaption}
\usepackage{lscape}
\usepackage{stmaryrd}
\usepackage{fancyhdr,graphicx,amsmath,amssymb}
\usepackage{graphicx,scalerel,stackengine}
\usepackage{float}
\usepackage[utf8]{inputenc}
\usepackage{booktabs}
\usepackage{dcolumn}
\usepackage{pdflscape}
\usepackage[linesnumbered,ruled,vlined]{algorithm2e}
\usepackage[mathscr]{euscript}
\usepackage{breqn}
\usepackage{cite}
\usepackage{amsmath,amssymb,amsfonts}
\usepackage{algorithmic}
\usepackage{graphicx}
\newcommand{\Lagr}{\mathcal{L}}
\usepackage{subcaption}
\usepackage{dirtytalk}
\usepackage{textcomp}
\usepackage{xcolor,colortbl}
\usepackage{mwe} 

\begin{document}
\title{A Robust and Scalable Attention Guided Deep Learning Framework for Movement Quality Assessment}

\author{Aditya Kanade$^{1}$,~Mansi~Sharma$^{1}$,~\textit{Member},~\textit{IEEE},~Manivannan~Muniyandi$^{2}$. 
\thanks{
$^1$The authors are with Department of Electrical Engineering, Indian Institute of Technology Madras, Chennai, India, 600036.

$^2$The author is with Department of Applied Mechanics, Indian Institute of Technology Madras, Chennai, India, 600036.

E-mail: \{ee20s086@smail,~mansisharma@ee\}.iitm.ac.in, mani@iitm.ac.in
}}

\maketitle

\begin{abstract}
Physical rehabilitation programs frequently begin with a brief stay in the hospital and continue with home-based rehabilitation. Lack of feedback on exercise correctness is a significant issue in home-based rehabilitation. Automated movement quality assessment (MQA) using skeletal movement data (hereafter referred to as skeletal data) collected via depth imaging devices can assist with home-based rehabilitation by providing the necessary quantitative feedback. This paper aims to use recent advances in deep learning to address the problem of MQA. Movement quality score generation is an essential component of MQA. We propose three novel skeletal data augmentation schemes. We show that using the proposed augmentations for generating movement quality scores result in significant performance boosts over existing methods. Finally, we propose a novel transformer based architecture for MQA. Four novel feature extractors are proposed and studied that allow the transformer network to operate on skeletal data. We show that adding the attention mechanism in the design of the proposed feature extractor allows the transformer network to pay attention to specific body parts that make a significant contribution towards executing a movement. We report an improvement in movement quality score prediction of 12\% on UI-PRMD dataset and 21\% on KIMORE dataset compared to the existing methods.
\end{abstract}

\IEEEpeerreviewmaketitle

\section{Introduction}
\label{sec:intro}

Brain stroke is a leading cause of severe long-term disability. Stroke reduces mobility in more than half of the survivors, especially senior citizens (65 years and above). A great effort has been directed towards improving the quality of life of stroke survivors with a wide range of technology \cite{technology-in-rehab}. However, only a few medical institutions are exploiting computer based rehabilitation tools \cite{home-rehab}. Literature surveys indicate that more than 90\% of rehabilitation sessions are carried out at home \cite{ninty-pc-rehab}. The patients either have to self-monitor or take help from the family members to monitor progress in the rehabilitation program. Since the home-based rehabilitation program is completely voluntary in nature, it often leads to low levels of patient adherence; resulting in prolonged post hospitalization recovery \cite{prolonged-1, prolonged-2}. Another issue with the home-based rehabilitation program is the lack of corrective feedback on movement quality and correctness. Some of the technological solutions available to the patient undergoing home-based rehabilitation are the robotic assistive systems \cite{robotic-assistance}, virtual reality and gaming interfaces \cite{gaming-interface}, and Kinect based assistance \cite{kinect-rehab}.

In their work, Liao et al. \cite{state-of-art} took a step in the direction of automated movement quality assessment from skeletal data. They proposed a deep learning framework for quantitative movement assessment; validated on the UI-PRMD dataset \cite{ui-prmd}. The first stage of their framework consists of a statistical model for the exercise generated by training a mixture of Gaussians on the correct performances of an exercise. They use the negative log-likelihood of the trained model to measure performance quality, termed the performance metric. A scoring function is defined, which maps the performance metric values into movement quality scores in the range [0, 1]. In the second stage, a hierarchical multiscale CNN-LSTM based architecture is employed to model this relationship between the movement data and the score. We list down a few shortcomings of their framework.

\begin{itemize}
    \item{In the first stage of their framework, the authors used an autoencoder to reduce the dimensionality of the movement data, which was used to train the GMM model. Autoencoders are prone to overfitting on the training data; without requisite regularization, the model will not learn good and general lower-dimensional representations \cite{denoising-ae}.}
    
    \item{Long Short Term Memory (LSTM), a recurrent neural network, is a significant building block in their model. LSTMs have been shown inferior in modeling very long temporal dependencies, they also suffer from poor scalability and longer training times \cite{transformer-orig}.}
    
\end{itemize}

In this article, we study advances in deep learning and propose a more robust and scalable framework for movement quality assessment, building on the work done by Liao et al. \cite{state-of-art}. The novel contributions of this paper are:

\begin{itemize}

    \item{ Three novel skeletal data augmentation schemes are proposed.}
    
    \item{A new training mechanism for the autoencoder network is proposed. We show that using the three proposed skeletal data augmentation leads to better outcomes in the existing performance score generation techniques.}
    
    \item{A novel transformer based architecture is proposed for movement quality assessment.}
    
    \item{Use of CNN based feature extractors is proposed to extract vector representation of temporal window slices, allowing the transformer network to operate on the skeletal data. We show that embedding the attention mechanism into feature extraction can allow the transformer network pay attention to specific body parts contributing to performance of an exercise.}

    \item{A comparative study is undertaken to choose the best feature extractor for the transformer model.}
    
\end{itemize}


\section{Proposed Architecture}
\label{sec:proposed-model}
Towards movement assessment and scoring, we begin our discussion by acquainting the reader with skeletal data collected on depth imaging devices such as Kinect and Vicon optical system. We then discuss three skeletal data augmentation techniques. We investigate application of skeletal data augmentation techniques to existing performance score generation techniques. Finally, a detailed study is carried out on the proposed Transformer architecture for predicting movement quality scores.

\subsection{Skeletal Data}
\label{sub:skeletal-data}
We consider movement data captured on a depth imaging system such as the marker-based Vicon optical tracker or the markerless Kinect system. These systems capture the movement data at a certain frame rate. At every frame, the system captures spatial information of several skeletal landmarks of the human body in 3-dimensional joint orientation and joint position. We consider joint orientation data captured by the system for our experiments due to invariance of the captured data to varying body structures. 
Considering an example system that tracks M skeletal landmarks on a human subject. Performance of a single repetition by a human subject of an exercise will generate data for $T$ frames, joint orientation information for all the $M$ landmarks will be captured at every frame by a vector $x^{(i)}, i \in [1...T]$ of dimension $D$, here $D = M \times 3$. Stacking these vectors for each frame results in a tensor of $X \in \mathbb{R}^{T \times D}$; a representative of the single repetition.

In our study, we consider two datasets; listed in Table \ref{tab:dataset}. The UI-PRMD dataset was collected using two devices, a Vicon optical tracker and a markerless Kinect tracker. The authors' Vicon configuration tracks 39 joint landmarks, whereas the Kinect-based system tracks 25 joint landmarks. Both Kinect and Vicon systems track joint position and orientation.
\begin{table}[ht]
    \centering
    \caption{Dataset Details}
\resizebox{\columnwidth}{!}{%
    \begin{tabular}{c c c c}
      \hline\hline
    Dataset &  No. of Participants & No. of Exercises & Depth Imaging System\\
      \hline
    UI-PRMD \cite{ui-prmd} &  10 & 10 &Vicon, Kinect V2\\
    KIMORE \cite{kimore} &  78 & 5 & Kinect V2\\
      \hline \hline
    \end{tabular}}
    \label{tab:dataset}
\end{table}
Data for the KIMORE dataset was captured on a Kinect camera. Additionally, performances by subjects are evaluated on a scale of $[0, 50]$ by five clinicians and are given as part of the KIMORE dataset.

\subsection{Skeletal Data Augmentation}
\label{sub:skeletal-data-aug}
Human motion displays a broad spectrum of variability. The data captured on the above-listed systems will always be limited in their sample size. Improving the generalization of machine learning models trained on this limited amount of data is of prime importance. Augmenting existing datasets by some form of mathematical manipulation is a popular method for improving the generalization capability of machine learning model \cite{augmentation-effectiveness}. We introduce three novel techniques of data augmentation for skeletal data.

\begin{itemize}
    \item {\textbf{Variable Pace of Exercise Performance} - The pace of exercise performance is highly variable. A machine learning model predicting a movement assessment score should depend on the movement's quality and not on the pace of performance. Increasing or decreasing the pace of exercise performance on already recorded movement data can be achieved by linear interpolation for decreasing the pace and downsampling for increasing the pace of performance.
}    
    \item {\textbf{Joint Occlusion} - Joint occlusion is a common phenomenon while collecting movement data. The movement data capturing system relies on skeletal landmarks to track body movements; complex body movements can result in the system losing track of the landmarks. The skeletal data is looked at through small windows of size \textit{h}, a small number of joints (\textit{n}) are randomly selected, and movement data recorded for these joints is set to zero within this window; simulating the condition of joint occlusion. Repeating movement data captured for all selected joints at the beginning of each window for the entire duration is another way to achieve this augmentation.}
    
    \item {\textbf{Movement Data Masking} - Input data masking is a popular technique of data augmentation in image classification. The skeletal data is looked at through small windows of size \textit{h}, entire movement data within this window is set to zero with some probability \textit{p} to achieve this augmentation.}
\end{itemize}

\begin{figure*}
\begin{subfigure}{0.245\textwidth}
\centering
\includegraphics[width=\linewidth]{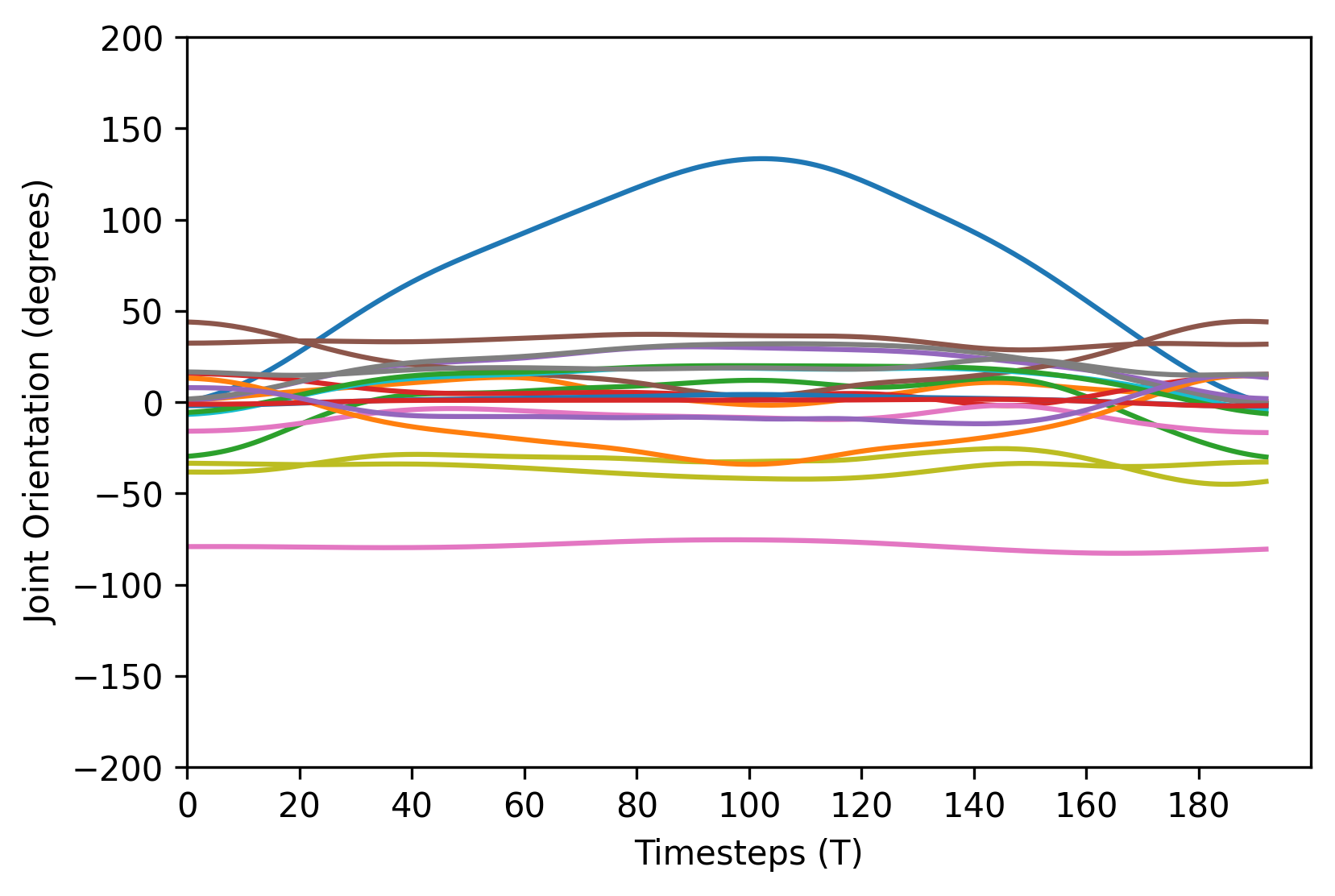}
\caption{}
\label{fig:subim2}
\end{subfigure}
\begin{subfigure}{0.245\textwidth}
\centering
\includegraphics[width=\linewidth]{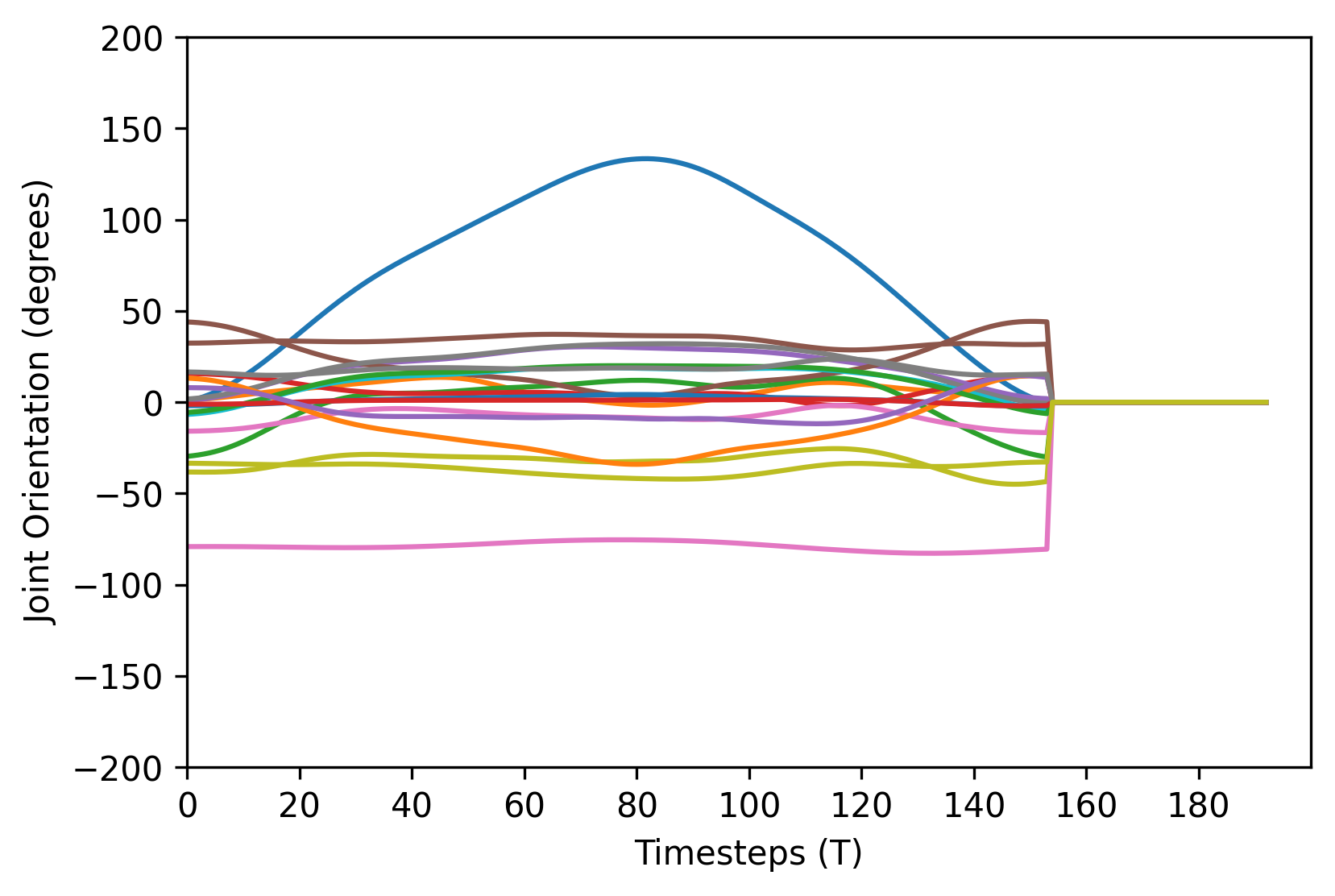}
\caption{}
\label{fig:subim2}
\end{subfigure}
\begin{subfigure}{0.245\textwidth}
\centering
\includegraphics[width=\linewidth]{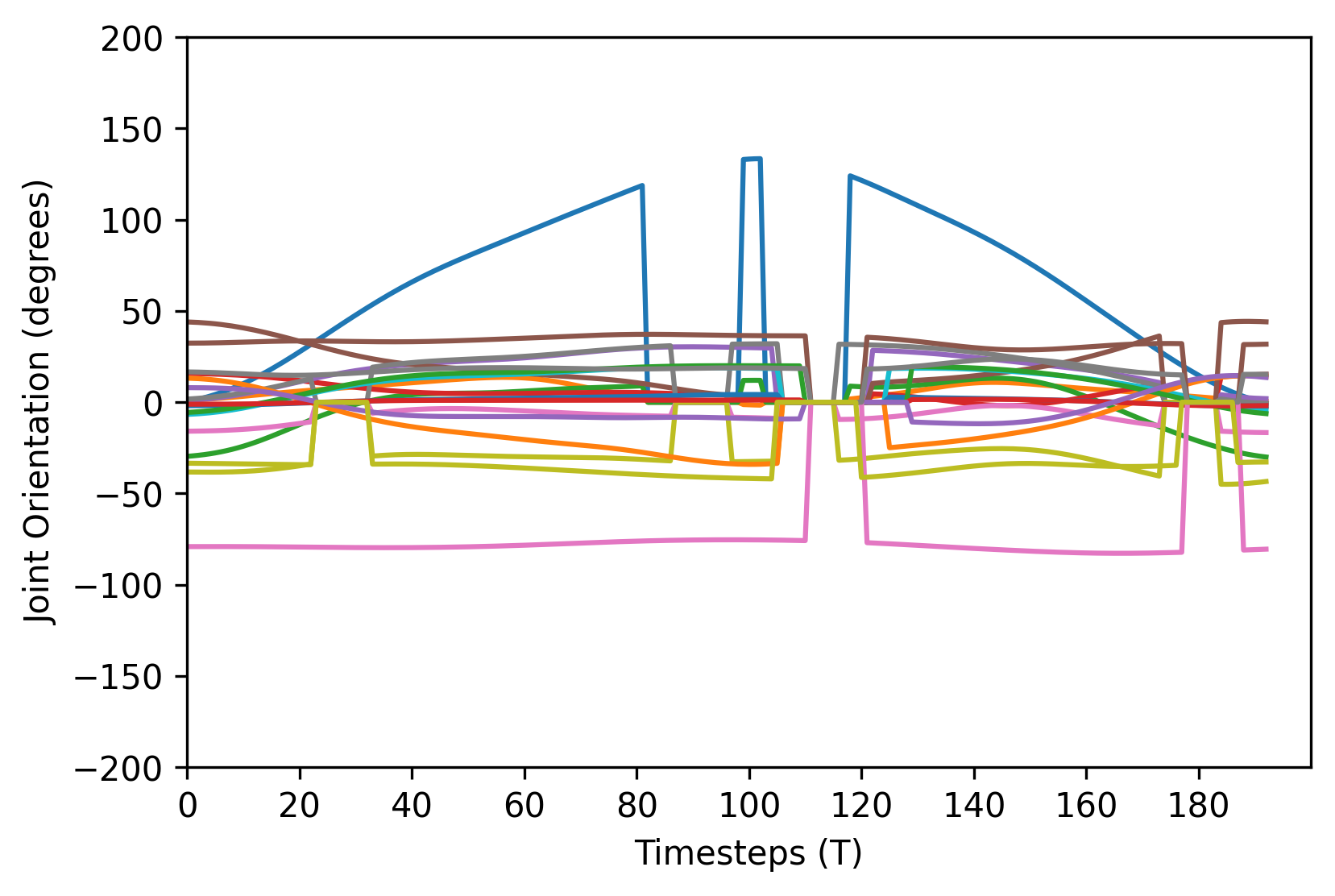} 
\caption{}
\label{fig:subim1}
\end{subfigure}
\begin{subfigure}{0.245\textwidth}
\centering
\includegraphics[width=\linewidth]{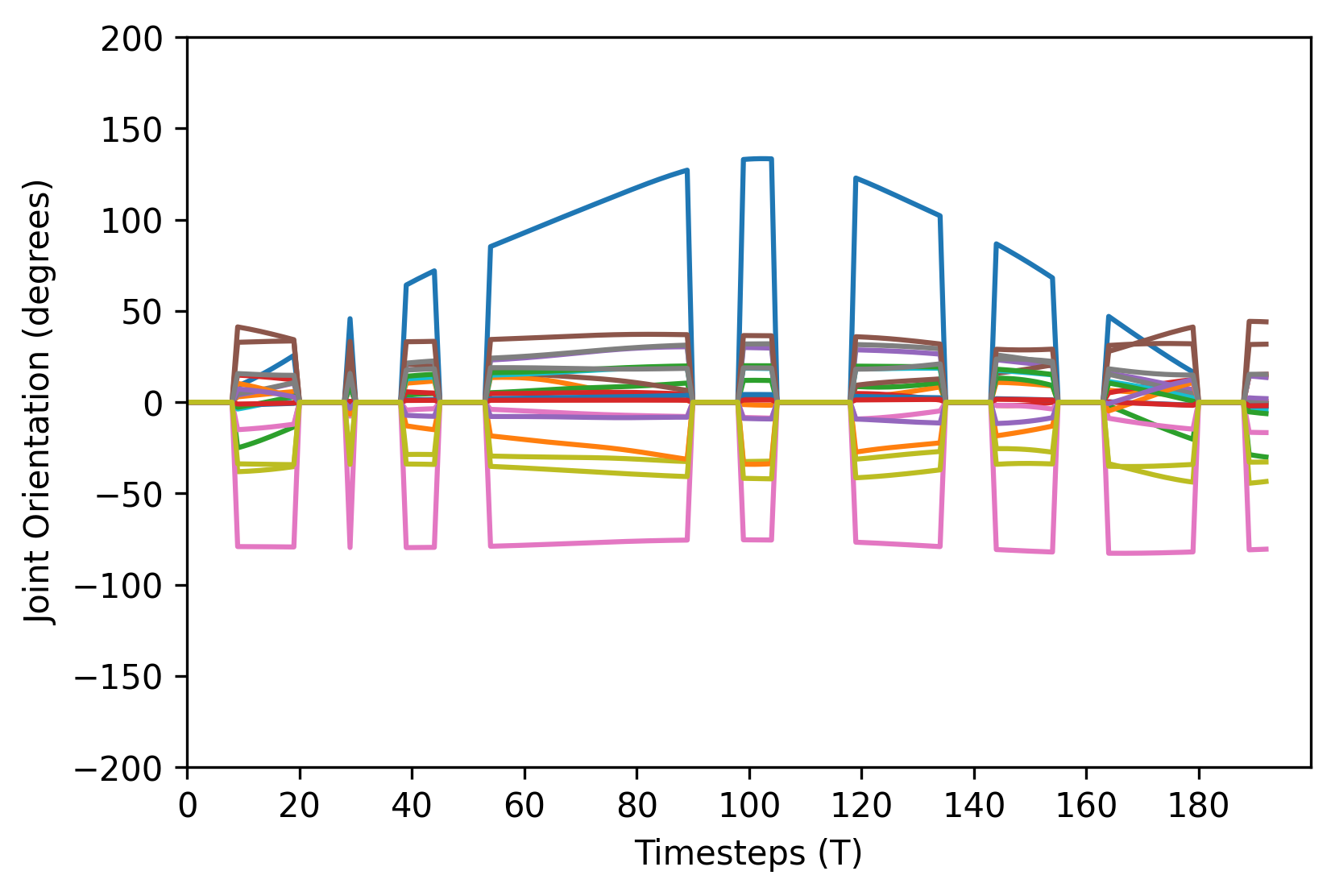} 
\caption{}
\label{fig:subim1}
\end{subfigure}
\caption{\textbf{Skeletal Data Augmentations - } Result of applying the three proposed skeletal data augmentations is shown here. We consider joint orientation data for six joints for better interpretability. The x-axis represents the number of timesteps (T) for which the data was recorded, while, the y-axis represents the orientation angle in degrees. \textbf{(a) Raw Skeletal Data - } Raw joint orientation data captured for six joints. Joint orientation data is represented in terms of three-dimensional Euler angles for each joint; resulting in 18 features per frame. \textbf{(b) Variable Pace of Exercise Performance - } The pace of performance is speeded up by a factor of 25\% by downsampling the raw skeletal data. \textbf{(c) Randomized Joint Occlusion - } The joints from frames are occluded by setting \textit{h} = 10 and \textit{n} = 2. \textbf{(d) Randomized Movement Data Masking - } The raw skeletal data is masked by setting \textit{h} = 10 and \textit{p}  = 0.2.}
\label{fig:augmentation}
\end{figure*}

Fig \ref{fig:augmentation} gives a graphical description of the discussed skeletal data augmentations.

\subsection{Movement Quality Score Generation}
\label{score-generation}
In a real-world home-based physical rehabilitation scenario, a quantitative measure of exercise performance can enable the patient to gauge their progress towards functional recovery. Using supervised regression, training a deep learning model for quantitative movement assessment is possible. However, such a model will require human movement data as input and an assessment score as output. Two ways can generate the assessment scores: 1) human expert-based assessment, 2) statistical model-based assessment. Each repetition of an exercise performed is scored by a trained clinician on a predefined scale in the human expert-based assessment. This method, however, presents a challenge in terms of consistency of movement assessment. The same motion can receive differing scores from different experts, reducing the reliability of such a scoring scheme. However, the statistical modeling method is different; here, the human bias does not affect the scoring. The model, however, is limited by the type of statistical model chosen to capture information of movement data for an exercise. 

Liao et al. \cite{state-of-art} proposed using a mixture of Gaussians to model an exercise. The model was trained a lower-dimensional projection of the data, to circumnavigate the curse of dimensionality. Their exploration found the autoencoder network to work best for this purpose. The authors explained their choice using a metric termed as the \textit{separation degree} for evaluating performances of various dimensionality reduction techniques. The authors explained the metric as follows \say{\textit{When applied to the values of the distance metrics, the separation degree indicates greater ability of the used metric to differentiate between correct and incorrect repetitions of an exercise}}. Here, by distance metric, the authors mean the statistical distance between a sample skeletal data recorded for an exercise and the trained exercise model. In this paper, we aim to improve the process of score generation, precisely the technique used for low dimensional projection of the movement data, and show the resulting improvements on the \textit{separation degree} metric. 

Research indicates that autoencoders tend to overfit the training data \cite{autoencoder-overfit}. Overfitting is problematic when lower-dimensional projection of the movement data from the autoencoder network is used to train the mixture of Gaussians since this model will only work well on the training data and will not generalize well. Learning good and general representations of the input movement data is essential. A large corpus of work indicates masked input modeling as a suitable choice for learning better representations in autoencoders \cite{masked-ae}. The random masking of the input ensures that a certain sense of the structure is built into the autoencoder model, allowing it to recover the original input from the masked input. We use the three novel skeletal data augmentation schemes proposed in Section \ref{sub:skeletal-data-aug} to train the autoencoder network. The new autoencoder training mechanism makes it a type of Denoising Autoencoder \cite{masked-ae}. We believe that the proposed training mechanism aids in the network learning better latent representations for the input movement data.

The autoencoder network consists of two components: 1) \textit{Encoder $f(.)$:} A deep neural network which models the lower dimensional latent representation of the input, 2) \textit{Decoder $g(.)$:} A deep neural network which projects the lower dimensional latent representation back into the input space. The encoder and decoder are jointly trained on the reconstruction loss $\Lagr_{reconst}(.)$ This loss is defined as the mean squared error between the reconstructed output and input at the encoder. A regularization factor $\Omega(\emph{f})$ is added to the reconstruction loss. The regularization is performed using $L_1$ norm of the encoder weights. This assists the proposed model to generalize better \cite{regularization}. Let $\Delta^{B} = \{X^{1}, X^{2}, ..., X^{B}\}$ represent a batch of B episodes of movement data for an exercise.   $\Bar{\Delta^{B}} = \{\Bar{X^{1}}, \Bar{X^{2}}, ..., \Bar{X^{B}}\}$ represents the augmented version of the batch. The neural network is trained using the following loss function.
\begin{equation}
    \Lagr_{total}(\Delta^{B}) = \Lagr_{reconst.}(\Delta^{B};\emph{g}(\emph{f}(\Bar{\Delta^{B}}))) + \lambda.\Omega(\emph{f})
\end{equation}

This training mechanism forces the network to learn the structure of movement data better, as the network is forced to predict the original movement data irrespective of the type of augmentation performed. 
\begin{figure}[h!]
\centering
  \centerline{\includegraphics[width=7cm, height=3.5cm]{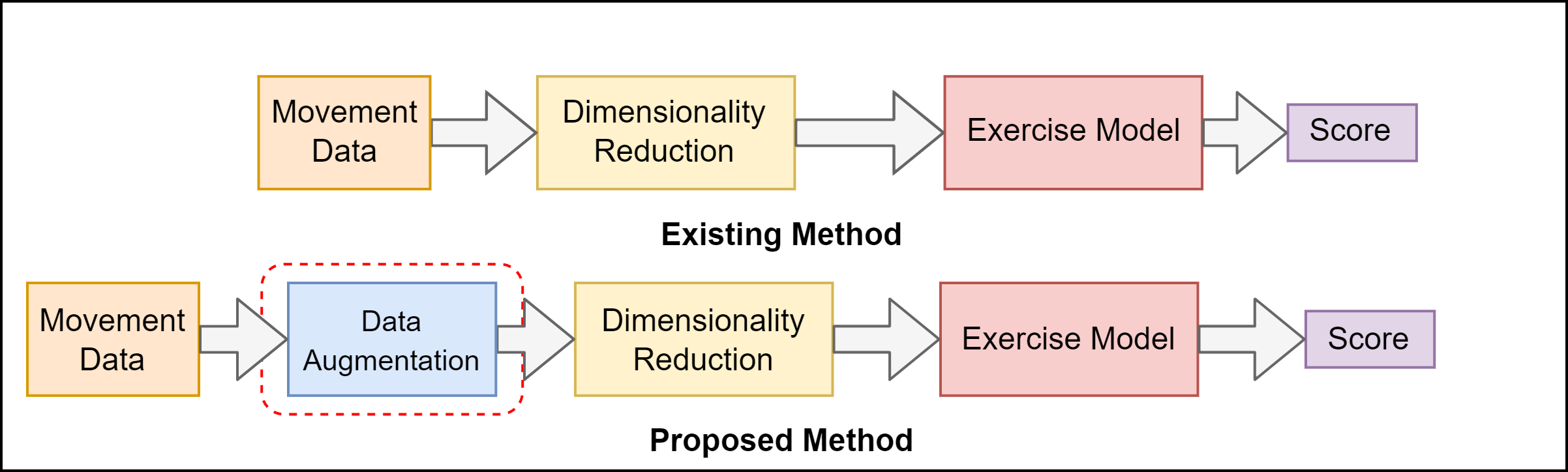}}
  \caption{\textbf{Score Generation Model - } The dotted red box highlights the proposed addition of the skeletal data augmentation in the existing performance score generation pipeline.}
\label{fig:score-gen}
\end{figure}

Fig~\ref{fig:score-gen} shows a block diagram of the score generation system in detail. These generated scores will be used to train the proposed transformer-based network to predict movement assessment scores. The quality of the model is evaluated using the \textit{separation degree} metric.

\subsection{Transformer Architecture for Rehabilitation Assessment}
\label{transformer}
Transformers have been consistently successful in tasks such as image classification, automatic speech recognition, natural language processing \cite{bert, vit, wave2vec2}. The transformer architecture is scalable, and a higher representation capacity can be obtained by increasing the number of layers in the architecture \cite{higher-representation}. Additionally, parallel execution is also possible since the transformer architecture is not recurrent. This architectural difference significantly reduces training time and allows for training on higher amounts of data. Training on more extensive data can pave the way for more complicated and scalable modeling of human movements as more movement data becomes available.

Movement data is harder to train on a transformer directly. For example, data for a single repetition of an exercise performed for 10 secs, captured on a Vicon system at 90fps will result in a tensor $X$ of size $\mathbb{R}^{900 \times117}$. Passing this tensor through a single attention block of the transformer will cost $900^2$ operations. Thus, passing the raw movement data input directly to the transformer does not seem reasonable. It leads to a quadratic increase in the number of parameters as the number of attention blocks increases.

Taking inspiration from the adaptation of the transformer architecture in the field of image classification and automatic speech recognition, we make use of an embedding layer between the raw movement data and the transformer for tokenizing the input data. The raw input $X \in \mathbb{R}^{T \times D}$ is split into N temporal window slices of size W each, such that $T = W \times N$, given by the set S, where each $s_i$ represents the $i^{th}$ temporal window.
\begin{equation}
S =  (s_1, s_2, s_3, ..., s_N), s_i \in \mathbb{R}^{W \times D}
\end{equation}
An embedding layer $E$ projects each temporal window into a latent subspace $Z$, where $E : s_i \longrightarrow z_i, i \in {1...N}$. This operation reduces the dimensionality to $\mathbb{R}^{K}$ such that each vector $z_i$ is a K dimensional vector representation of the $i^{th}$ temporal window $s^i$. The transformer architecture does not have a way to capture positional information of the input tokens by design. We use the technique described in \cite{transformer-orig} to add positional information to the input. Each vector $z_i$ is added with a positional embedding vector $p_i \in \mathbb{R}^{K}$. The set $Z_p$ is the result of this operation.
\begin{equation}
Z_p =  (z_1 + p_1, z_2 + p_2, z_3 + p_3, ..., z_N + p_N), z_i \in \mathbb{R}^{K}
\end{equation}

\begin{figure}[h!]
  \centerline{\includegraphics[width=6cm, height=10cm]{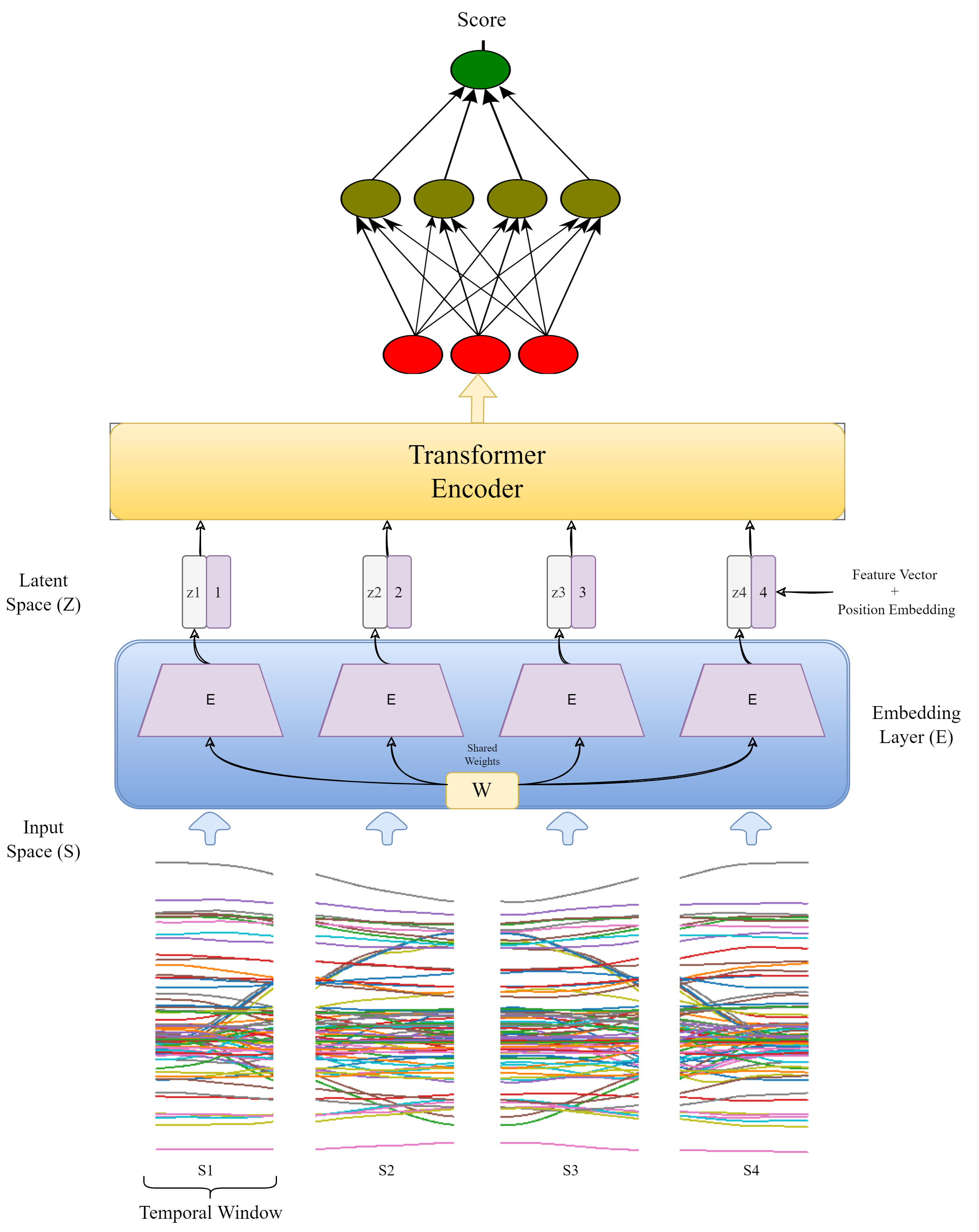}}
\caption{Network Architecture}
\label{fig:net-arch}
\end{figure}
The set $Z_p$ is passed on to the lowest encoder layer of the transformer encoder. \textit{Note:} we only use the transformer encoder in our experiment. The attention map generated at the topmost layer of the transformer encoder is projected through a series of dense layers, which finally predict a score value for the exercise. The network uses the binary cross entropy between the predicted and the actual assessment score to jointly learn the parameters of the transformer encoder and the embedding layer. The architecture of this network is shown in Fig \ref{fig:net-arch}. We additionally experiment with the structure of the embedding layer to select the best performing feature extractor. A comparison between feature extractors is reported in the results section. We consider four feature extractors for the embedding layer in our study: 1) MLP Feature Extractor, 2) CNN Feature Extractor, 3) Hierarchical Feature Extractor (HFE), and 4) Attention guided Hierarchical Feature Extractor (HFE-A). The MLP and CNN-based feature extractors are simple in structure; the former consists of fully connected neurons layers, while the latter consists of a stack of one-dimensional convolutional filters.
\begin{figure}[h!]
  \centerline{\includegraphics[width=7cm, height=6cm]{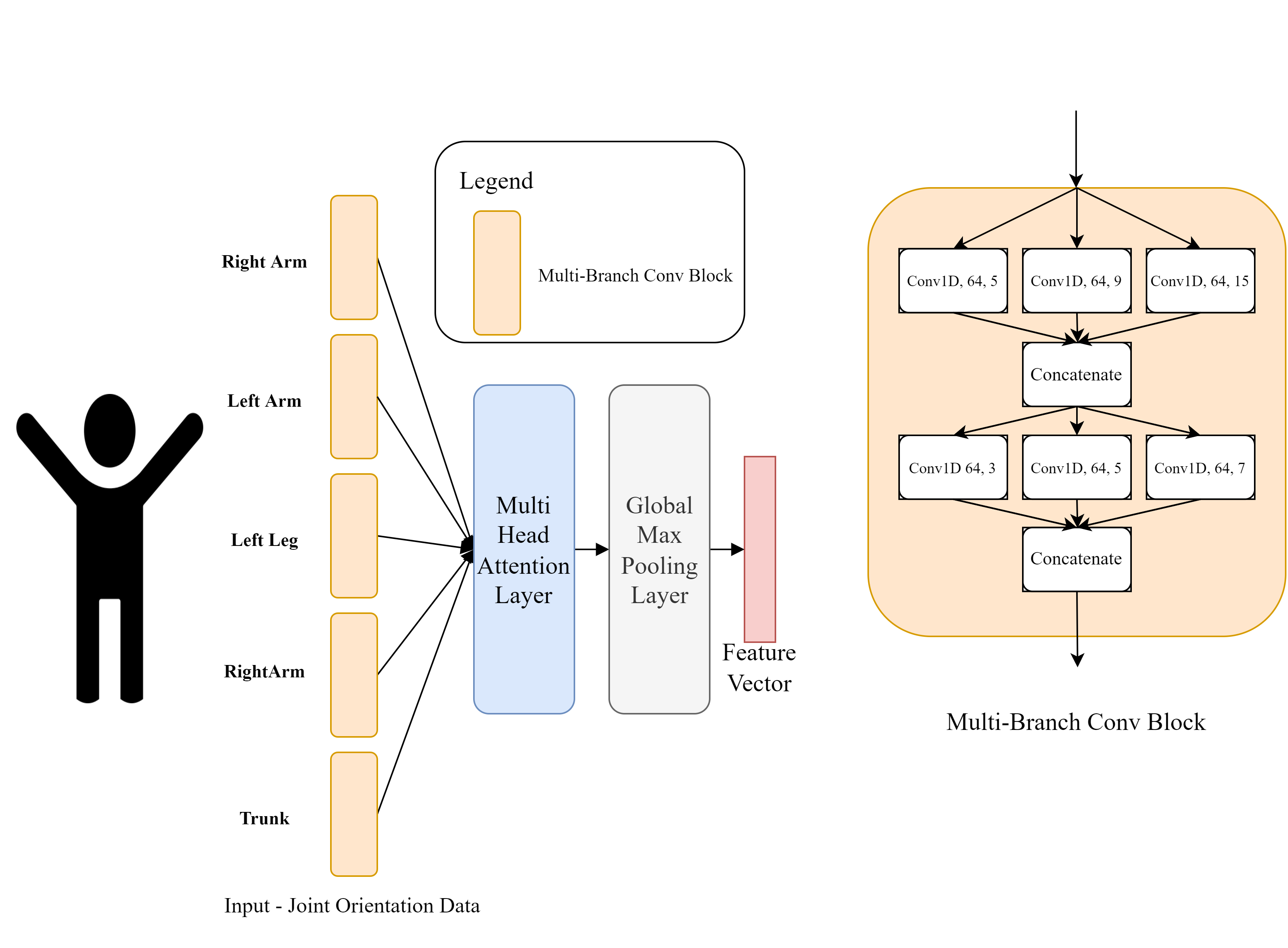}}
  \caption{Attention guided Hierarchical Feature Extractor}
\label{fig:hierarchical}
\end{figure}
The previous body of work has shown better performance exploiting the hierarchical structure of the movement data for action recognition \cite{hierarchical, state-of-art}. We adopt this structure and propose a novel feature extractor called the Hierarchical Feature Extractor (HFE). The feature extractor is designed to exploit the spatial characteristics of human movements by dedicating sub-networks for processing joint displacements of individual body parts. We propose another novel feature extractor termed the Attention guided Hierarchical Feature Extractor (HFE-A). This feature extractor consists of a multi-head attention block after the hierarchical feature extractor. This block computes a soft mixture of the features extracted by the individual sub-networks. For our architecture, we use five heads in the multi-head attention block. The resulting computation of the multi-head attention block is passed through a global max pooling block which selects the most dominant features. At the end of this pooling block, the feature vector becomes the vector representation of the temporal widow slice of the input movement data.

\section{Results and Discussions}

\subsection{Movement Quality Score Generation Performance}
\label{score}
The scores for movement data in UI-PRMD dataset are generated based on the discussion in Section \ref{score-generation}. Augmentation schemes discussed in section \ref{sub:skeletal-data-aug} are used; augmentations are applied for every batch. This ensures that the network is never trained on the original version of the data. At every iteration, the network sees a modified version of data ensuring that overfitting doesn't occur on the training data. Table \ref{tab:separation-degree} shows the results on \textit{separation degree} metric.

We see a significant improvement in the separation degree metric performance for the proposed novel training mechanism, with an increase of 1.5 percent in \textit{within-subject} and a 18 percent increase in the \textit{between-subject} category. The autoencoder model with the novel training mechanism proposed in Section \ref{score-generation} has to predict the original movement data irrespective of the augmentation applied at the input side. The resulting lower dimensional latent structure that the model discovers is more general since it displays a certain invariance to the augmentation applied; since it has to reconstruct this latent representation back to the original input. A better representation allows more capacity to differentiate between correct and incorrect movements, leading to greater \textit{separation degree}, as shown by the results.

 \begin{center}
 \begin{table}[ht]
 \centering
\caption{\textbf{Separation Degree (\textit{SD})}- shows that the use of skeletal data augmentations for training an autoencoder results in better separation between the correct and incorrect movements.}
\label{tab:separation-degree}
\resizebox{\columnwidth}{!}{%
\begin{tabular}{|c|cc|cc|}
\hline
& \multicolumn{2}{c|}{Within Subject \textit{(SD)}} & \multicolumn{2}{c|}{Between Subject \textit{(SD)}}\\ \cline{2-3} \cline{4-5}
&  \multicolumn{1}{c|}{Deep Rehab \cite{state-of-art}} &  \multicolumn{1}{c|}{Proposed}  & \multicolumn{1}{c|}{Deep Rehab \cite{state-of-art}} &  \multicolumn{1}{c|}{Proposed}\\ 

\hline
\hline
E1-E10  &  0.511   & 0.518    & 0.402   & 0.478\\
\hline
   E1   &  0.376          & \textbf{0.540}  & 0.288          & \textbf{0.444}\\
   E2   &  \textbf{0.642} & 0.571           & 0.404          & \textbf{0.606}\\
   E3   &  \textbf{0.510} & 0.352           & \textbf{0.423} & 0.281\\
   E4   &  \textbf{0.585} & 0.571           & 0.338          & \textbf{0.499}\\
   E5   &  0.489          & \textbf{0.511}  & \textbf{0.428} & 0.374\\
   E6   &  0.394          & \textbf{0.478}  & 0.387          & \textbf{0.552}\\
   E7   &  \textbf{0.617} & 0.550           & 0.525          & \textbf{0.597}\\
   E8   &  0.525          & \textbf{0.535}  & 0.414          & \textbf{0.507}\\
   E9   &  0.461          & \textbf{0.497}  & 0.339          & \textbf{0.370}\\
   E10  &  0.510          & \textbf{0.571}  & 0.480          & \textbf{0.552}\\
\hline
\end{tabular}}
\end{table}%
\end{center}

\subsection{Transformer Model Performance}
The proposed model is trained on the movement data from the UI-PRMD dataset. The raw joint orientation data of the skeletal landmarks are used as input to the proposed network. The scores for exercises are generated using techniques described in section \ref{score-generation}. The network is designed to predict the movement quality score in a supervised regression setting. The model was implemented on an HP desktop computer with an i7 processor, 16GB RAM, and an NVIDIA-2080Ti GPU card. A separate model is trained for each of the ten exercises in the UI-PRMD dataset. We report the model performance in terms of the average absolute deviation between the ground truth movement quality scores and the network prediction; these values are averaged over five runs for generating the results. The network is trained on a 0.8/0.2 train/validation split. The model is trained using the Adam optimizer, setting the learning rate to 0.0005 \cite{adam}. Early stopping is used to avoid overfitting the training data. A patience value of 100 epochs is set to monitor the validation loss. The network is trained on binary cross-entropy loss between the predicted score and the ground truth scores. An extensive grid search was carried out for selecting hyperparameters and fine-tuning the transformer encoder and the embedding layer. Following is a list of the best configuration we found for the proposed model.
\begin{itemize}
    \item Temporal Window ($W$) Size - $\textbf{40}$
    \item Feature Vector ($z_i$) Size - $\textbf{256}$
    \item Number of Heads in Transformer Encoder Block- $\textbf{4}$
    \item Number of Transformer Encoder Blocks - $\textbf{2}$
\end{itemize}

To evaluate the performance of various feature extractors for the embedding layer, we perform an ablation study. The results on Deep Squat exercise (E1) in the UI-PRMD dataset are displayed in Table \ref{tab:feat-ext-results}. 

 \begin{table}[H]
\centering
\begin{tabular}[t]{ccc}
\hline
Embeddor Type & MAE & Time per Batch\\
\hline
\hline
   MLP               &  0.0180          & 30ms  \\
   CNN               &  0.0215          & 30ms  \\
   HFE      &  0.0200          & 205ms \\
   HFE-A  &  \textbf{0.0175} & 215ms \\
\hline
\end{tabular}
\caption{\textbf{Comparative Study of Feature Extractors for the Embedding Layer}
- The study was carried out on Deep Squat Exercise (E1) from the UI-PRMD dataset. We report Average Absolute Deviation for all four types of the proposed feature extractors. We also report the time taken by each network per batch.}
\label{tab:feat-ext-results}
\end{table}%

\begin{figure}[h!]
  \centerline{\includegraphics[width=8cm, height=3cm]{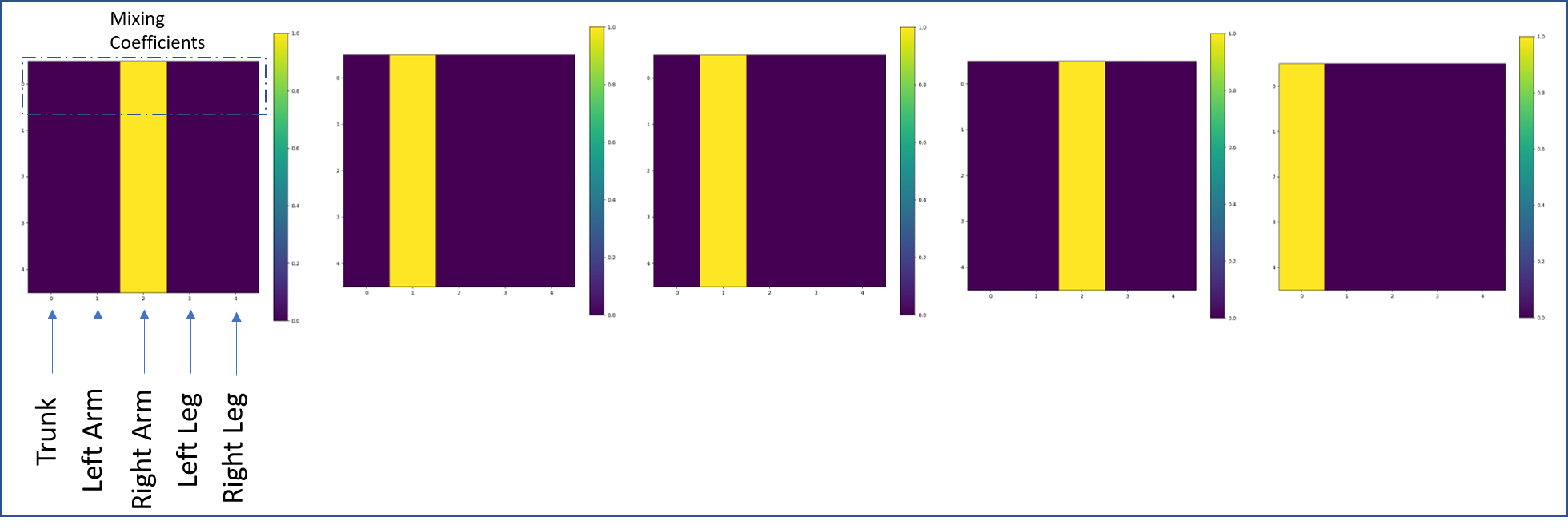}}
  \caption{\textbf{Attention Score Map - } The figure shows the attention map computed by five heads of the multi-head attention block for Standing Shoulder Abduction exercise (E7) in UI-PRMD dataset. Each row of the figure indicates mixing coefficients for vector output from each sub-network. We see that the network pays attention to upper-body sub-networks since the exercise is primarily an upper-body exercise.}
\label{fig:attention-score}
\end{figure}

We see that the proposed HFE-A works best between the four proposed feature extractors for the embedding layer. Fig \ref{fig:attention-score} shows attention map generated at the multi-head attention layer of HFE-A. The attention map was generated by training the proposed transformer network with HFE-A for the embedding layer on Standing Shoulder Abduction (E7) exercise. The exercise is primarily an upper-body exercise. The results from Fig \ref{fig:attention-score} support our intuitive assumption that the multi-head attention block introduced in the design of HFE-A allows the proposed transformer network to attend specific body parts that contribute significantly to an exercise while predicting movement quality score. Finally, it is important to note that the simple MLP based feature extractor also yields comparable results to HFE-A while being computationally very efficient. Accordingly, we find that both these feature extractors can be considered based on the trade-off between accuracy and computational efficiency.

\begin{table}[ht]
\centering
\caption{Average Absolute Deviation on Movement Quality Scores for the UI-PRMD Dataset}
\label{tab:main-results}
\begin{tabular}[t]{p{1.3cm}p{1.3cm}p{1.3cm}p{1.3cm}p{1.3cm}}
\hline
&Proposed Model & Deep Rehab \cite{state-of-art} & Deep CNN \cite{state-of-art} & Deep LSTM \cite{state-of-art}\\
\hline
\hline
E1-E10 & \textbf{0.0365} & 0.0415  & 0.0456 & 0.0418 \\
E1  & \textbf{0.0175} & 0.0229  & 0.0286 & 0.0230 \\
E2  & \textbf{0.0321} & 0.0399  & 0.0438 & 0.0363 \\
E3  & \textbf{0.0369} & 0.0441  & 0.0394 & 0.0414 \\
E4  & \textbf{0.0342} & 0.0367  & 0.0344 & 0.0350 \\
E5  & \textbf{0.0321} & 0.0362  & 0.0384 & 0.0336 \\
E6  & \textbf{0.0331} & 0.0414  & 0.0424 & 0.0373 \\
E7  & \textbf{0.0544} & 0.0565  & 0.0751 & 0.0598 \\
E8  & \textbf{0.0379} & 0.0418  & 0.0448 & 0.0448 \\
E9  & \textbf{0.0252} & 0.0283  & 0.0364 & 0.0402 \\
E10 & \textbf{0.0613} & 0.0672  & 0.0727 & 0.0668 \\
\hline
\end{tabular}
\end{table}%

We now compare the performance of proposed transformer model to current state-of-the-art deep learning models for movement assessment. We are aware of only one such study which applies deep learning to the field of movement quality assessment; hence we compare our results with two additional baseline models, Deep CNN and Deep LSTM. The model parameters and architecture for the baseline models were selected based on the details listed in \cite{state-of-art} for a fair comparison. Table \ref{tab:main-results} lists the results of comparative analysis between the models.


The results demonstrate over 12\% improvement over the current state-of-the-art method and the baseline models on UI-PRMD dataset. We also show results on KIMORE dataset, which was collected on the Kinect V2 system. We followed the same 0.8/0.2 train/validation split for this dataset. The results of the score predictions are shown in Table \ref{tab:kimore-results}. 

 \begin{table}[ht]
\centering
\caption{Average Absolute Deviation on Movement Quality Scores for the KIMORE Dataset}
\label{tab:kimore-results}
\begin{tabular}[t]{p{1.3cm}p{1.3cm}p{1.3cm}p{1.3cm}p{1.3cm}}
\hline
&Proposed Model & Deep Rehab \cite{state-of-art} & Deep CNN \cite{state-of-art} & Deep LSTM \cite{state-of-art}\\
\hline
\hline
E1-E5 & \textbf{0.1093} & 0.1390  & 0.1282 & 0.1500 \\
E1  & \textbf{0.0826} & 0.1456  & 0.1346 & 0.1257 \\
E2  & 0.1624 & \textbf{0.1429}  & 0.1448 & 0.1670 \\
E3  & \textbf{0.0767} & 0.1399  & 0.1205 & 0.1661 \\
E4  & \textbf{0.1097} & 0.1344  & 0.1192 & 0.1442 \\
E5  & \textbf{0.1153} & 0.1326  & 0.1219 & 0.1464 \\
\hline
\end{tabular}
\end{table}%

The results demonstrate over 21\% improvement over the current state-of-the-art method and the baseline models on the KIMORE dataset. Both these results show superior performance by the proposed model. At this point, it is important to note that finding the right hyperparameters for the Transformer based architecture is more difficult than CNN-LSTM based architectures. However, the increased accuracy, scalability and speed of the Transformer-based network architecture compensates for this. Currently, the field of movement assessment lacks a large-scale dataset containing a large number of diverse examples of various exercises prescribed for physical rehabilitation. We believe that the proposed model can aid in scaling to much larger datasets as a result of the Transformer based architecture. 

\section{Conclusion}
The aim of this paper was to address the problem of movement quality assessment using deep learning. In this paper, we introduced three novel data augmentation schemes for human skeletal data. We show that using the proposed augmentation schemes for performance score generation results in an improvement of over 18\% in \textit{between-subject} category and 1.5\% in the \textit{within-subject} category on the \textit{separation degree} metric. The enhanced results are due to more general latent representations learned by the autoencoder network, allowing the exercise model better differentiation between correct and incorrect movements. We further proposed a novel Transformer based architecture for movement quality assessment and introduced four novel feature extractors for the embedding layer, aiding the transformer to operate on the continuous skeletal data. We compare the performance of the proposed transformer network on all the proposed feature extractors and show that the attention guided hierarchical feature extractor (HFE-A) and the MLP feature extractor give the best results. We finally compare the proposed network against current state-of-the-art methods and show improvement of over 12\% on the UI-PRMD dataset and 21\% on the KIMORE dataset. Improvements in the movement quality scores are aided with enhanced explainability of the score prediction process due to the addition of the attention layer. 


In this paper, we have explored the technique of masking and pace manipulation as a way of data augmentation. However, further research should be directed towards more augmentation schemes for skeletal data. Better augmentation schemes can aid the application of newer techniques such as self-supervised learning to the field of skeletal data. The attention mechanism in the proposed HFE-A feature extractor is a beginning towards the application of attention mechanism in the field of movement assessment. Further research needs to be undertaken for searching for even more powerful ways to apply the attention mechanism for skeletal data.

\section*{Acknowledgment}

The scientific efforts leading to the results reported in this paper have been carried out under the supervision of Dr. Mansi Sharma while working as an INSPIRE Hosted Faculty, IIT Madras.

\bibliographystyle{ieeetr}
\footnotesize
\bibliography{root.bib}
\end{document}